\def\year{2022}\relax
\documentclass[letterpaper]{article} 
\usepackage{aaai22}  
\usepackage{times}  
\usepackage{helvet}  
\usepackage{courier}  
\usepackage[hyphens]{url}  
\usepackage{amsmath}
\usepackage{amsfonts}
\usepackage{amsthm}
\usepackage{graphicx}
\usepackage{float}
\usepackage{stfloats}
\usepackage{multirow}
\usepackage{array}
\usepackage{graphicx} 
\usepackage{natbib}  
\usepackage{caption} 
\DeclareCaptionStyle{ruled}{labelfont=normalfont,labelsep=colon,strut=off} 
\usepackage{algorithm}
\usepackage{algorithmic}

\newcolumntype{P}[1]{>{\centering\arraybackslash}p{#1}}

\usepackage{newfloat}
\usepackage{listings}
\floatstyle{ruled}
\newfloat{listing}{tb}{lst}{}
\floatname{listing}{Listing}
\begin{document}
%
\title{Attacking Deep Reinforcement Learning-Based Traffic Signal Control Systems with Colluding Vehicles} 
\author{Ao Qu,\textsuperscript{1}
Yihong Tang,\textsuperscript{2}
Wei Ma,\textsuperscript{3}\\
\textsuperscript{1}{Vanderbilt University}\\
\textsuperscript{2}{Beijing University of Posts and Telecommunications}\\
\textsuperscript{3}{The Hong Kong Polytechnic University}\\
ao.qu@vanderbilt.edu,
tyh@bupt.edu.cn, 
wei.w.ma@polyu.edu.hk}
\maketitle
\begin{abstract}
\begin{quote}
The rapid advancements of Internet of Things (IoT) and artificial intelligence (AI) have catalyzed the development of adaptive traffic signal control systems (ATCS) for smart cities. In particular, deep reinforcement learning (DRL) methods produce the state-of-the-art performance and have great potentials for practical applications. In the existing DRL-based ATCS, the controlled signals collect traffic state information from nearby vehicles, and then optimal actions ({\em e.g.,} switching phases) can be determined based on the collected information. The DRL models fully ``trust'' that vehicles are sending the true information to the signals, making the ATCS vulnerable to adversarial attacks with falsified information.
In view of this, this paper first time formulates a novel task in which a group of vehicles can cooperatively send falsified information to ``cheat'' DRL-based ATCS in order to save their total travel time. To solve the proposed task, we develop \textsc{CollusionVeh}, a generic and effective vehicle-colluding framework composed of a road situation encoder, a vehicle interpreter, and a communication mechanism. We employ our method to attack established DRL-based ATCS and demonstrate that the total travel time for the colluding vehicles can be significantly reduced with a reasonable number of learning episodes, and the colluding effect will decrease if the number of colluding vehicles increases. Additionally, insights and suggestions for the real-world deployment of DRL-based ATCS are provided. The research outcomes could help improve the reliability and robustness of the ATCS and better protect the smart mobility systems.

\end{quote}
\end{abstract}

\section{Introduction}
\begin{figure}
    \centering
    \includegraphics[width=1\columnwidth]{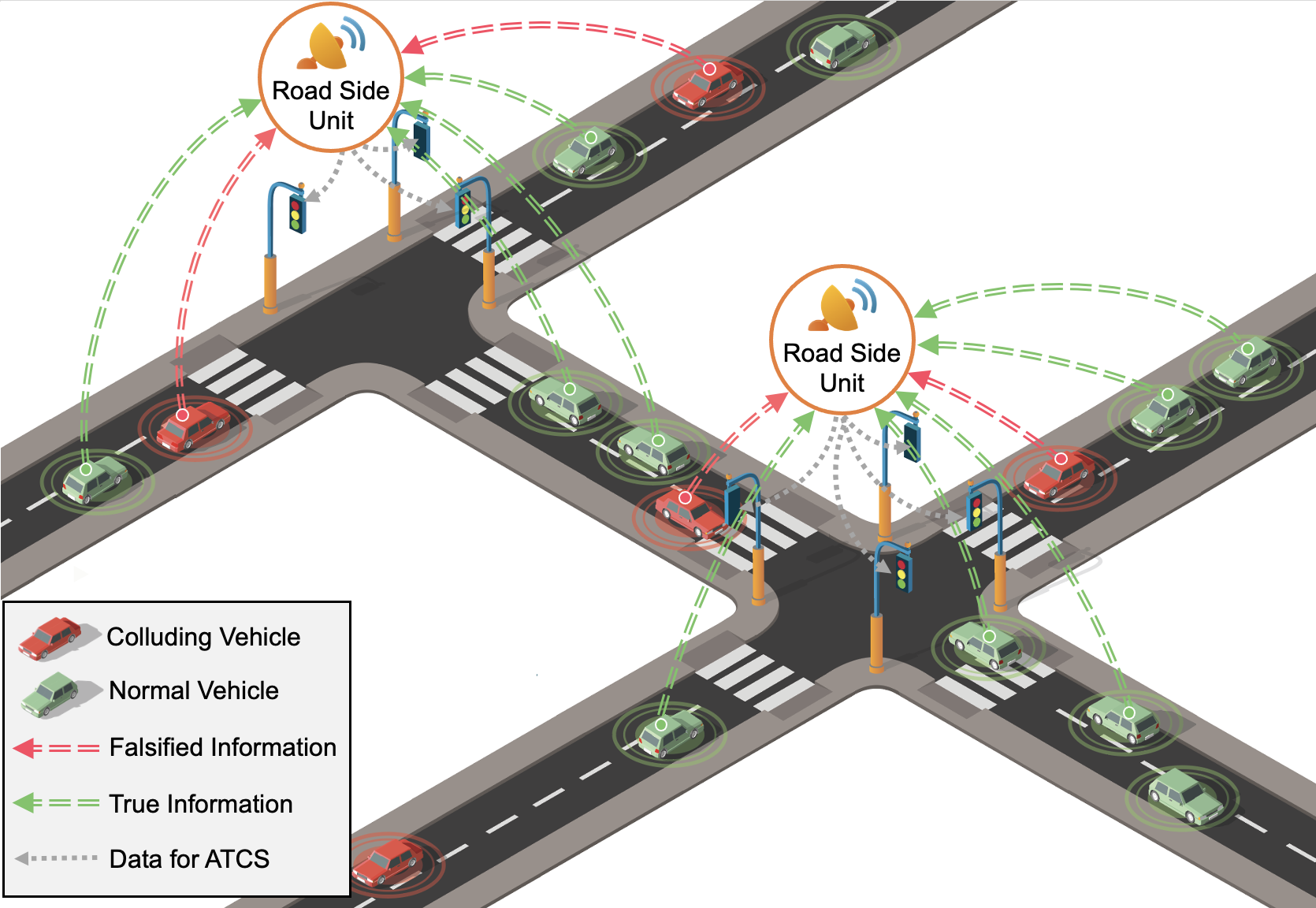}
    \caption{Attacking ATCS with colluding vehicles.}
    \label{fig:problem}
\end{figure} 
Being able to accommodate changing traffic patterns and mitigate congestion by adjusting the duration of each signal phase, Adaptive Traffic Signal Control Systems (ATCS) have been widely appreciated and adopted as an alternative approach to the traditional fixed-time signal control system. For example, Alibaba has deployed a traffic management system called ``City Brain'' in more than 20 cities in China as well as in other countries bringing demonstrated ease in congestion \cite{zme_science_2021}. These industrial level deployments in real-world provide great incentives for the development of ATCS. 
The recent success in deep reinforcement learning (DRL) has catalyzed increasing interest in developing DRL-based signal control algorithms which learn by trial-and-error to develop dynamic policy. The DRL-based ATCS respond to the real-time traffic condition, making traffic signals adaptive to the changing world. In academia, the KDD Cup competition themed as ``City Brain Challenge'' has attracted 1100+ teams over the globe to design coordination algorithms for city-scale traffic signal control \cite{city_brain_challenge}. At the same time, governments have been investing heavily in smart infrastructure ({\em e.g.,} Biden's \$2 trillion infrastructure plan \cite{the_white_house_2021}), bringing more possibilities for technologies such as Vehicle-to-Vehicle (V2V) and Vehicle-to-Infrastructure (V2X) in the near future. With these advancements, we have good reasons to expect massive deployments of DRL-powered ATCS in the near future with more efficient information communication enabled by connected vehicles and smart infrastructure. 

However, as the functioning of such systems involves receiving signals for the collection of traffic data, the benefits may come with potential risks. Specifically, cyber-attacks have been witnessed on many cities' digital networks including camera systems and computer network \cite{freed_2021,cranley_2020}. Consequently, as traffic signals are getting connected with vehicles, such connectivity may open new doors to potential cyber attacks as well. According to a recent report published by UC Berkeley, smart traffic lights are identified as one of the top 3 most vulnerable smart city technologies \cite{UCBerkelyReport}. Meanwhile, as many works have pointed out, DRL algorithms tend to exhibit great vulnerabilities under malicious attacks, making the adoption of DRL-based ATCS a challenge \cite{behzadan2017vulnerability,behzadan2018faults,DBLP:journals/corr/abs-1905-10615,ilahi2021challenges}.

Few studies investigate the robustness of traffic signal control under cyber attacks and all existing works only examine potential harms caused by attacks \cite{doi:10.1177/0361198118756885,Chen2018ExposingCA}. Indeed, vehicles can obtain benefits such as reduced travel time by attacking the traffic signal systems ({\em e.g.}, sending falsified information). To our best knowledge, no previous work studies related issues, although this is an realistic scenario as such attacks can be practiced easily with strong motivations. Given the massive amount of cost associated with the implementation of ATCS, it is important to thoroughly comprehend the limitations and potential instabilities of the current DRL-based ATCS models in order to work towards more robust design.

In view of this, we first time formulate a novel problem where a small fraction of vehicles can form a collusion aiming to reduce their total travel time by cooperatively sending falsified information. The problem is considered to take place in a connected environment where each vehicle can communicate with other vehicles and send signals to ATCS via road side units to indicate its presence. In previous works, the target ATCS's policy is known but this is not realistic in practice especially when traffic signals are controlled by DRL agents\cite{doi:10.1177/0361198118756885}. In our problem, ATCS is controlled by trained DRL agents whose policies are hidden from all vehicles. In order to infer knowledge about an obscure policy, we attempt to solve this problem by treating the colluding vehicles as a multi-agent system where reinforcement learning algorithm is applied to help them find solution. An illustration of this problem is shown in Fig. \ref{fig:problem} where red vehicles are those in the collusion group. Several factors are worth considering when dealing with this novel problem. One is that although the overall strategy for colluding vehicles is to cooperate, competitions for resources happen whenever more than one colluding vehicle come to the same intersection seeking different traffic signal phases. Another factor is the spatio-temporal variation in traffic patterns which may affect ATCS's policy as well. For example, a signal trained at a constantly busy intersection may tend to perform differently than a signal that has never seen heavy traffic. 

To model the above-mentioned factors, we propose a generic DRL-based framework \textsc{CollusionVeh}. The framework first leverages parameter sharing techniques to generate embeddings for traffic scenarios, allowing vehicles to together explore the spatio-temporal variations and different traffic scenarios. Then, we assign each vehicle an unique network to interpret the produced embeddings. In the end, we design a communication module that enables agents to exchange information for better coordination.
In summary, our paper makes the following contributions:
\begin{itemize}
    \item For the first time, we formulate a novel problem in which a vehicle collusion group can cooperatively attack a black-boxed DRL-based ATCS with the common goal of reducing total travel time.
    \item We propose \textsc{CollusionVeh}, which, to our best knowledge, is the first DRL-based vehicle-attacking framework composed by three extendable key modules that are effective at both capturing global features and communicating for better coordination. 
    \item By conducting comprehensive ablation studies and sensitivity tests, we identify the factors that contribute to successful attacks. Based on these results, we provide insights and suggestions on how to improve the current ATSC systems.
\end{itemize}

\section{Related Work}
\subsubsection{Reinforcement Learning Based Adaptive Traffic Signal Control}
For a traffic signal control system with $N$ intersections, an intersection $i$ is often treated as an agent that can take action ({\em e.g.}, switching to next signal phase) $a_{S_i}^t$ based on its most recent observation $s_{S_i}^t$ of the surrounding traffic ({\em e.g.}, the number of vehicles on each adjacent road). Depending on the problem setting, ATSC can be considered as either a single-agent task where the agent aims to learn an uniform policy for all signals \cite{6082823,casas2017deep} or a multi-agent task where each agent acts as a signal to learn individual policies with a common goal of improving traffic shared by all agents \cite{8569301,10.1145/3219819.3220096,Chen_Wei_Xu_Zheng_Yang_Xiong_Xu_Li_2020}. In the context of IoT, communication strategies are also designed for agents to improve the robustness \cite{chu2019multi,Wei_2019,10.5555/3398761.3399082,wang2020stmarl,Xu_Wang_Wang_Jia_Lu_2021}.
\subsubsection{Attacking Adaptive Traffic Signal Control Systems}
DRL based models are known to be vulnerable to adversarial perturbations. However, the research area of the adversarial attacks against the DRL applications is still largely untouched, as the majority of the studies focus on perturbing observations in toy environments such as in Atari Games \cite{pmlr-v48-mniha16}. Specifically, the adversary has the ability to alter an agent's action through adding perturbations to their observations \cite{10.1007/978-3-319-62416-7_19,pmlr-v78-dosovitskiy17a,DBLP:journals/corr/abs-1905-10615,DBLP:journals/corr/abs-2003-08938,DBLP:journals/corr/abs-2101-08452} proposes a novel algorithm in which adversary can create natural observations that act as adversarial inputs to make the agent follow desired policy, and these work prove the possibility of attacking DRL policies under different settings.\\
In the context of attacking traffic signal control systems, \cite{doi:10.1177/0361198118756885} proposes an optimization-based method to test the vulnerability of ATCS by sending falsified information to maximize the network-wide delay, in which ATCS's policies are known in advance. However, ATCS's policy should be seen as a black box in real world. To address this gap, we apply DRL-based adversaries to attack DRL-based ATCS.

\subsubsection{Vehicular Ad Hoc Network}
Vehicular ad hoc network (VANET) is a subclass of mobile ad hoc networks (MANETs) that comprises self-organizing vehicles as mobile nodes, which  was first mentioned and introduced in \cite{toh2001ad}. VANET includes vehicle-to-vehicle (V2V) and vehicle-to-infrastructure (V2I) communications \cite{isaac2008secure}, and this kind of mechanism has been used for enhancing safety ({\em e.g.}, collision avoidance, traffic optimization, etc.) and comfort ({\em e.g.}, toll/parking payment, locating fuel stations, etc.) \cite{khan2017certificate}. 
In terms of DRL, the concept of VANET has been introduced in multi-agent autonomous \& connected vehicle setting. Recent work has demonstrated the effectiveness of dynamic graph information sharing mechanisms \cite{gunarathna2019dynamic,chen2021graph,wangmulti}.

\section{Preliminaries}
\subsubsection{Reinforcement Learning}
RL models an agent that maximizes its rewards when interacting with an environment without having any prior knowledge. The interaction process can be represented as a fully/locally observable Markov Decision Process (MDP) \cite{bellman1957markovian}. For each interaction with the MDP, an agent observes the state $s_{t}\in\mathcal{S}$ and performs an action $a_{t}\in\mathcal{A}$ according to a policy $\pi(a{\mid}s)$, where $\mathcal{S}$ is the state space and $\mathcal{A}$ is the action space. Then the environment yields an immediate reward $r_{t}=\mathcal{R}(s_{t},a_{t},s_{t+1})$ and transits to the next state $s_{t+1}$, which is taken over $s_{t+1} \sim \mathcal{T}\left(\cdot \mid s_{t}, a_{t}\right)$. In a finite MDP, the expected cumulative reward starting from state $s$, following policy $\pi$, taking action $a$, is defined as a $\mathcal{Q}$-function: $\mathcal{Q}^{\pi}(s, a)=\mathbb{E}_{\pi}\left[\sum_{\tau=t}^{\infty} \gamma^{\tau-t} r_{\tau} \mid s_{t}=s, a_{t}=a\right]$, where $\gamma \in [0,1)$ is the discount parameter indicating the weight of future rewards. The value function is obtained by summing the $\mathcal{Q}$-function over the action space: $\mathcal{V}^{\pi}(s)=\sum_{a \in \mathcal{A}} \pi(a \mid s) \mathcal{Q}^{\pi}(s, a)$. The objective of an agent is to find the optimal policy $\pi^{*}$ that can maximize the expectation of rewards, so that the optimal value function is: $\mathcal{V}^{*}(s)=\max \mathcal{V}^{\pi}(\mathrm{s}), \forall s \in \mathcal{S}$. From the optimal value function we can derive the optimal policy: $\pi^{*}(s)=\underset{\pi}{\arg \max } \mathcal{V}^{\pi}(\mathrm{s}), \forall s \in \mathcal{S}$.

\begin{figure*}[t!]
    \centering
    \includegraphics[width=.9\textwidth]{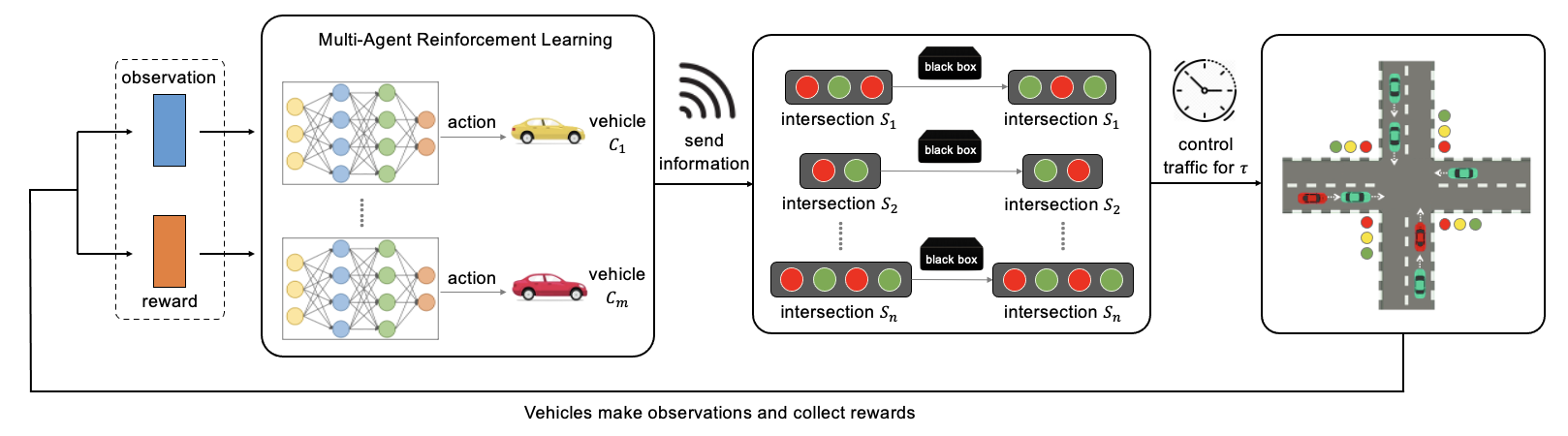}
    \caption{A flowchart of the proposed problem for colluding vehicles attacking ATCS.}
    \label{fig:flowchart}
\end{figure*}
\subsubsection{Proximal Policy Optimization} RL algorithms can be divided into policy-based methods and value-based methods. In value-based RL, the RL agent updates a value function at each iteration, in policy-based algorithms, for each iteration, policy is updated through policy gradients \cite{sutton2018reinforcement}. Traditional policy gradient algorithms are time-consuming due to sample methods and does not guarantee good convergence. Proximal Policy Optimization (PPO) \cite{schulman2017proximal} is a state-of-the-art on-policy algorithm which is simpler to implement and has better sample complexity. To resolve the issue of traditional policy gradient algorithm, PPO adopts two mechanisms for better performance and stability. 

First, PPO formulates policy gradients using advantage-function: $\mathcal{A}^{\pi}(s, a)=\mathcal{Q}^{\pi}\left(s, a)-\mathcal{V}^{\pi}(s\right)$. Given a state $s_{t}$, performing an action $a_{t}$, the advantage-function measures relative advantage compared to other actions. Secondly, PPO updates the parameters within a certain trust region to ensure using the following objective function that the deviation from the previous policy is relatively small: $J^{\theta^{k}}(\theta)=\sum\limits_{(s_{t},a_{t})} \min (w, \operatorname{clip}(w, 1-\varepsilon, 1+\varepsilon))
    A^{\theta^{k}}\left(s_{t}, a_{t}\right)$, where $w=\frac{p_{\theta}\left(a_{t} \mid s_{t}\right)}{p_{\theta^{k}}\left(a_{t} \mid s_{t}\right)}$. In our work, We use the Actor-Critic approach for our PPO agents. Actor-Critic uses two estimators: Actor guides agents' actions based on policy and Critic evaluates the action. The probability distribution for choosing actions is updated in favor of actions that perform better than critic's evaluation.

\section{Problem Statement}
In this section, we formulate a novel problem concerning the security and robustness of RL-based ATCS by allowing vehicles to attack the trained ATCS with falsified information. A step-by-step flowchart of this problem is presented in Fig. \ref{fig:flowchart}. The environment is a road network with $N$ intersections where the traffic signal $S_i$ at intersection $i$ is controlled by a DRL-agent whose policy is unknown to any vehicle. The traffic signals have been trained to set its signal phase to optimize traffic flow. Whenever an agent controlling traffic signal $i$ is asked to produce the next action $a_{S_i}^t$ based on its current observation $o_{S_i}^t$, the selected signal phase will last for a duration $\tau$. Importantly, the observation $o_{S_i}^t$ is obtained by querying all the running vehicles $\textbf{V}$ on nearby road network. We further assume a small group of vehicles $\textbf{C}$ acting in collusion to send falsified information to traffic signals with the common goal of reducing their total trip duration. Each vehicle $C_i$ in $\textbf{C}$ is also controlled by a DRL-agent. At time step $t$, vehicle $C_i$ takes action $a_{C_i}^t$ based on its partial observation $o_{C_i}^t$ of the environment. During training, $C_i$ learns to find the optimal policy $\pi_{C_i}$ that maximizes the cumulative reward $r$.

\section{Methodology}
\begin{figure*}[h]
    \centering
    \includegraphics[width=.76\textwidth]{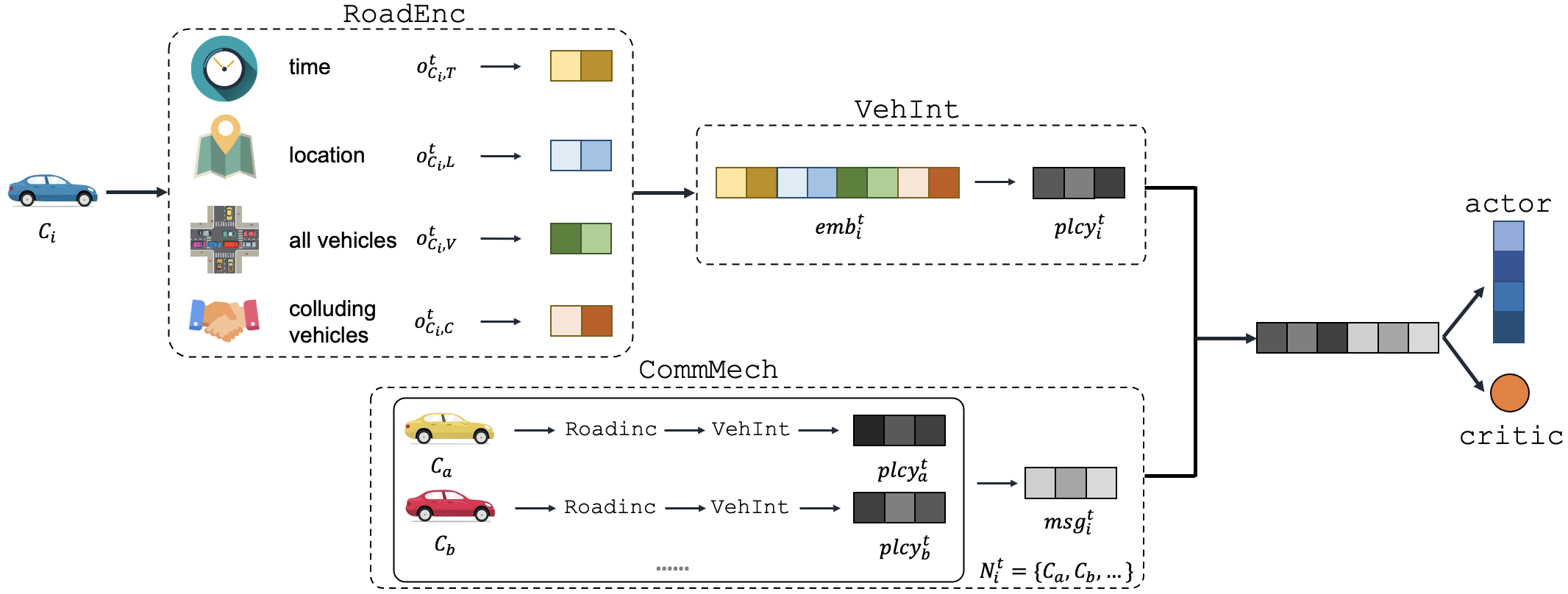}
    \caption{An illustration of the vehicle colluding framework \textsc{CollusionVeh}.}
    \label{fig:framework}
\end{figure*}

In this section, we first present the design of observation, action, and reward for each agent (eg. a colluding vehicle) which employs typical Proximal Policy Optimization (PPO) algorithm for training. Then, we propose a generic vehicle-colluding framework, \textsc{CollusionVeh}, to effectively help vehicles choose the optimal action based on its observation. The model is composed of a road situation encoder, a vehicle interpreter, and a communication mechanism which together capture the spatio-temporal dynamics of traffic and leverage parameter sharing and communication techniques. A detailed illustration of the proposed framework is shown in Fig. \ref{fig:framework}.  It is worth noting that the proposed model assumes the realization of fully connected vehicle scenario where agents can accomplish the learning together and send real-time information to partners. In order to handle scenarios where these capabilities are constrained, we also present a few variations of the proposed framework.

\subsection{Agent Design} 
Without knowing the specific policies of the ATCS in charge, it is important to formulate an effective and practical design of observation, action, and reward for our agents so that they can obtain a comprehensive understanding of the environment and develop inductive reasoning to achieve their goal. Since, to our best knowledge, such setting has never been introduced before, we will also discuss the rationality and feasibility of each design in order to facilitate further studies in this topic.
\subsubsection{Observation} At time step $t$, an agent $C_i$ is able to gain partial information about the environment as its current observation $o_{C_i}^t$. In \textsc{CollusionVeh}, a single observation consists of the following components: current time $\&$ location and traffic around the upcoming intersection. In particular, time can be implemented as the hour in a day in practice. In our study, we first divide the total duration into intervals with the same length (i.e. $\{T_1, T_2, \dots, T_k\}$) and then use one-hot encoding to represent any timestamp. Similarly, we use the upcoming intersection, represented as another one-hot vector, to describe location information. The traffic information is divided into two parts. One includes the general vehicle counts for all lanes approaching the upcoming intersection and the other one contains colluding vehicle counts constructed in a similar way. Therefore, one observation made by agents can be expressed as $o_{C_i}^t = [o_{C_i, T}^t, o_{C_i, L}^t, o_{C_i, V}^t, o_{C_i, C}^t]$, referring to time, location, and two types of traffic information respectively.\\
The first two components are included because traffic patterns are known to exhibit complex spatio-temporal variations. For example, agents may need to adopt a more aggressive policy during morning peaks or when approaching to a notoriously congested intersection. Most of the ATCS determine signal phase based on the surrounding traffic so it would be helpful for vehicles to obtain similar information. In particular, to avoid unnecessary competition and facilitate coordination, agents also get information about the surrounding colluding vehicles. 
\subsubsection{Action}
After receiving an observation $o_{C_i}^t$, the agent can send a falsified data indicating the presence of more than one vehicles in order to attract more "attention" from the traffic signal in charge. The action space $\textbf{A}$ for each agent is defined as a set of possible numbers that each agent can report and at time step $t$, each agent can select an action $a_{C_i}^t \in \textbf{A}$ as the number of vehicles it will pretend to be. \\
When V2I technology is applied, traffic signals make observation about the nearby traffic by receiving signals from vehicles indicating their presence. From a cybersecurity point of view, such signals can be verified via V2I certification management but, as a member of a collusion group, a vehicle might carry several certificates and therefore can pretend to be more than one vehicle.
\subsubsection{Reward}
With the goal of minimizing total travel time for all colluding vehicles, we define the reward for each agent as the negative average waiting time over all running agents since last actions are executed. Therefore, when agents attempt to maximize their reward, their waiting time will be reduced.  If we denote the set of agents that are running in the environment as \textbf{$C_{R}$} and for $C_i \in$ \textbf{$C_{R}$}, we denote its accumulated waiting time at time $t$ as $\textit{wait}_{C_i}^t$, then the immediate reward at time $t$ is defined as,
$$r_{t, i} = -\frac{1}{|C_{R}|} \sum_{i\in C_{R}} \left(\textit{wait}_{C_i}^t - \textit{wait}_{C_i}^{t-\tau}\right).$$

\subsection{Core Modules}
The proposed framework, \textsc{CollusionVeh}, is composed of three modules: a road situation encoder, a vehicle interpreter, and a communication mechanism. The idea is that the road situation encoder is able to generate a general representation for an arbitrary traffic scenario. Then, each agent has an independent vehicle interpreter which is able to interpret the previously generated traffic representation for its own use. In the end, agents inform their nearby partners via a communication mechanism for better coordination. 
\subsubsection{Road Situation Encoder}
A noticeable limitation exhibited by most current reinforcement learning algorithms is sample inefficiency. This barrier becomes even more significant in multi-agent reinforcement learning tasks as the high sample complexity makes the computation intractable. In order to enhance sample efficiency, we design a universal road situation encoder $\texttt{RoadEnc}$ which is trained on data collected from all agents so that it can effectively generate an informative representation of the traffic scenario. Previously, we have determined four types of features as the observation for some agent i, denoted as $[o_{C_i, T}^t, o_{C_i, L}^t, o_{C_i, V}^t, o_{C_i, C}^t]$. Then, $\texttt{RoadEnc}$ is a global encoding mechanism shared by all agents. Let \texttt{emb} be the output of \texttt{RoadEnc}, then we have
$$\texttt{emb}_i^t = \texttt{RoadEnc}\left(o_{C_i, T}^t, o_{C_i, L}^t, o_{C_i, V}^t, o_{C_i, C}^t\right).$$
In our study, given the independence between four features (eg. time feature does not tell anything abut location), we employ four separate global embedding mechanisms each of which is implemented as a multilayer perceptron (\textsc{MLP}). Consequently, \texttt{RoadEnc} is implemented as
\begin{multline*}
\texttt{RoadEnc}(o_{C_i, T}^t, o_{C_i, L}^t, o_{C_i, V}^t, o_{C_i, C}^t) = [\textsc{MLP}^T(o_{C_i, T}^t)\Vert \\ \textsc{MLP}^L(o_{C_i, L}^t) \Vert \textsc{MLP}^V(o_{C_i, V}^t) \Vert \textsc{MLP}^C(o_{C_i, C}^t)],
\end{multline*}
where $\Vert$ is the concatenation operation.

\subsubsection{Vehicle Interpreter}
Having its own origin and destination (OD) and preferred path, each vehicle should have a different interpretation of a given traffic scenario. For example, a congestion on a north-south road may not affect the south-north traffic even at the same intersection.  
Therefore, for each agent, we introduce another module, Vehicle Interpreter (\texttt{VehInt}), in order to understand the road situation embedding that is trained to comprehensively describe the upcoming traffic. For agent $i$, the output from this step, denoted as $plcy_i^t$, takes the following form:
$$plcy_i^t = \texttt{VehInt}(emb_i^t).$$
For simplicity, this module is implemented as an agent-specific \textsc{MLP}. 
\subsubsection{Communication Mechanism}
When multiple agents approach the same intersection and compete for the traffic signal phase in their favor, it is important for these agents to communicate and figure out the best coordination plan. An effective communication mechanism (\texttt{CommMech}) should be able to summarize neighboring policies into a message that contributes to the final decision process. For agent $i$, let $\textbf{N}_i$ denote the other agents approaching the same intersection, we have
$$msg_{i}^t = \texttt{CommMech}(\{plcy_i^t | i\in \textbf{N}_i\})$$
The main challenge is that $\textbf{N}_i$, typically seen as a vehicular ad-hoc network (VANET), is fast-changing and it is hard to determine $|\textbf{N}_i|$. A graph attention network (GAT) based model, \textsc{MARL-CAVG}, has been proposed in \cite{wangmulti} to simulate the VANET formulated by connected vehicles but it assumes a fixed number of running vehicles. In our setting, vehicles have different starting and ending time which makes \textsc{MARL-CAVG} inapplicable. For the sake of simplicity, we apply an \textsc{MLP} to the average of all neighboring policies to generate the message as follows: 
$$\texttt{CommMech}\left(\{plcy_j^t | j\in \textbf{N}_i\}\right) = \textsc{MLP}\left(\frac{1}{|\textbf{N}_i|} \sum_{j\in \textbf{N}_i} plcy_j^t\right).$$

\subsection{Actor Critic}
As we have introduced, our agents learn their policies using PPO algorithm which is based on the actor-critic approach. For each agent, its actor network and critic network share all but the last layer in order to reduce both time and space complexity. \\
Specifically, the actor network (\texttt{actor}) is implemented as
$$\texttt{actor}_i(o_{C_i}^t) = \sigma\left(\textsc{MLP}(plcy_i^t \Vert msg_i^t)\right),$$
where $\sigma$ is the softmax function and the output space has the same dimension with the agent's action space to estimate the probability of each action. The critic network (\texttt{critic}) is implemented similarly with the output dimension equal to one, approximating $V^{\pi}(s_{C_i}^t)$.

\section{Experiments}
We conduct numerical experiments on SUMO, a state-of-the-art microscopic traffic simulation software that allows users to control both traffic signals and vehicles. The target ATCS is trained using \textsc{MA2C} proposed by \cite{chu2019multi}, which is one of the best performing DRL-based ATCS algorithms. Most DRL-based ATCS algorithms have similar designs in terms of the agent's observation, action, and reward definitions, but \textsc{MA2C} is the only algorithm that takes account of both communication between neighboring intersections and spatial discount factors, an analogue of the discount factor in DRL in the context of spatial data, which greatly boost its robustness \cite{DBLP:journals/corr/MousaviS0H17,DBLP:journals/corr/abs-1905-04722,GONG2019100020,Chen_Wei_Xu_Zheng_Yang_Xiong_Xu_Li_2020}. The Monaco city traffic network is used for our experiment due to its large variety of intersection types: among 30 signalized intersections in total, 11 are two-phase, 4 are three-phase, 10 are four phase, 1 is five-phase, and the reset 4 are six-phase. The robustness of target ATCS policy and complexity of the road network allow us to fully investigate the properties of \textsc{CollusionVeh}. 

\subsection{Main Results}
\renewcommand*{\arraystretch}{1.15}
\begin{table*}[h]
\caption{Comparison with baseline models and ablation study (unit: seconds).} 
\label{table:main_results}
\resizebox{\textwidth}{!}{%
\begin{tabular}{@{\extracolsep{4pt}}lccccc@{}}
\hline
\multirow{2}{*}{}                                & \multirow{2}{*}{Reward} & \multicolumn{2}{c}{Colluding Vehicles} & \multicolumn{2}{c}{Other Vehicles} \\ \cline{3-4} \cline{5-6} 
                                                 &                         & Travel Time Avg   & Waiting Time Avg   & Travel Time Avg & Waiting Time Avg \\ \hline
\textsc{All One}                                          & -482.0±(88.63)          & 83.26±(1.97)      & 15.33±(2.99)       & 121.84±(0.21)   & 25.31±(0.19)     \\ 
\textsc{All Five}                                         & -147.21±(11.3)          & 72.49±(2.06)      & 4.6±(0.34)         & 131.2±(11.83)   & 34.51±(10.94)    \\ 
\textsc{All Ten}                                          & -61.97±(16.57)          & 69.84±(1.94)      & 1.95±(0.54)        & 134.58±(14.0)   & 37.59±(12.8)     \\ 
\textsc{Random}                                           & -128.02±(16.41)         & 70.99±(1.98)      & 3.11±(0.65)        & 131.04±(8.28)   & 34.69±(8.08)     \\ \hline
\texttt{VehInt}                                & -53.29±(16.43)          & 69.55±(2.08)      & 1.68±(0.36)        & 130.01±(18.56)  & 32.86±(17.22)    \\ 
\texttt{RoadEnc} (w/o spatial-temporal features) + \texttt{VehInt} & -45.72±(12.48)          & 69.54±(2.34)      & 1.58±(0.81)        & 125.26±(10.34)  & 29.23±(9.45)     \\ 
\texttt{RoadEnc} + \texttt{VehInt}                                 & -41.94±(12.47)          & 69.14±(2.08)      & 1.24±(0.41)        & 132.6±(12.51)   & 35.8±(11.83)     \\ \hline
\textbf{\texttt{RoadEnc} + \texttt{VehInt} + \texttt{CommMech} (\textsc{CollusionVeh})} & \textbf{-37.8±(9.13)} & \textbf{69.07±(1.84)} & \textbf{1.15±(0.23)} & \textbf{138.1±(13.66)} & \textbf{40.93±(12.6)} \\ \hline
\end{tabular}%
}
\end{table*}
For the primary experiment, we consider the scenario where 493 vehicles are running on the Monaco city road network with a random sample of 30 vehicles selected to form a collusion group. Specific routes are generated with randomness approximating a time variant traffic flow distribution. Each episode of training contains a 300 timestep simulation of the traffic flow during which most vehicles are able to complete their trips. 1,000,000 steps (~3000 episodes) of training are performed for each experiment. During training, each colluding vehicle agent has an action space of size 11 indicating that they can report as up to 10 vehicles instead of 1. Five metrics are employed to evaluate the experiment results. In addition to rewards as defined in previous section, we also measure the average travel time and average waiting time for both colluding and non-colluding vehicles. To address the stochasticity caused by random sampling process, we run each model with 5 random seeds and calculate the average and standard deviation across all 5 experiments. Since this problem is introduced for the first time, we define the following four rule-based attacking models as our benchmark.\\
\textbf{\textsc{All One}, \textsc{All Five}, and \textsc{All Ten}}: In \textsc{All One} model, every agent is honest to the traffic signals and reports itself as just one vehicle. \textsc{All Ten} is the greedy approach such that all vehicles always select the maximum possible action. \textsc{All Five} is defined similarly.\\
\textbf{\textsc{Random}}: Whenever an agent is able to send signals to the upcoming traffic light, it chooses a random number from 0 to 10 as its action.
We present the comparison between \textsc{CollusionVeh} and the four baseline models in Table \ref{table:main_results} (row 1-4 and 8). Not surprisingly, when agents take \textsc{All Ten} approach, they earn significantly greater rewards than \textsc{All One}. However, as we have discussed, the naive greedy strategy cannot handle coopetition, namely, the mix of cooperation and competition. Clearly, our proposed model, \textsc{CollusionVeh}, successfully solves this issue as the average cumulative reward rises up to $-37.8$ from $-61.97$ achieved by \textsc{All Ten} and \textsc{CollusionVeh} consistently beats all four baseline methods across all metrics. For the actual trips, the proposed framework reduces the total waiting time by ~92.5\% as compared to \textsc{All One} and by ~41.0\% as compared to \textsc{All Ten}. Meanwhile, there is an ~62.7\% increase in non-colluding vehicles' average waiting time, making them spend ~13.3\% more time on travel. \\

\subsection{Ablation Study}
\begin{figure}[h]
    \centering
    \includegraphics[width=.9\columnwidth]{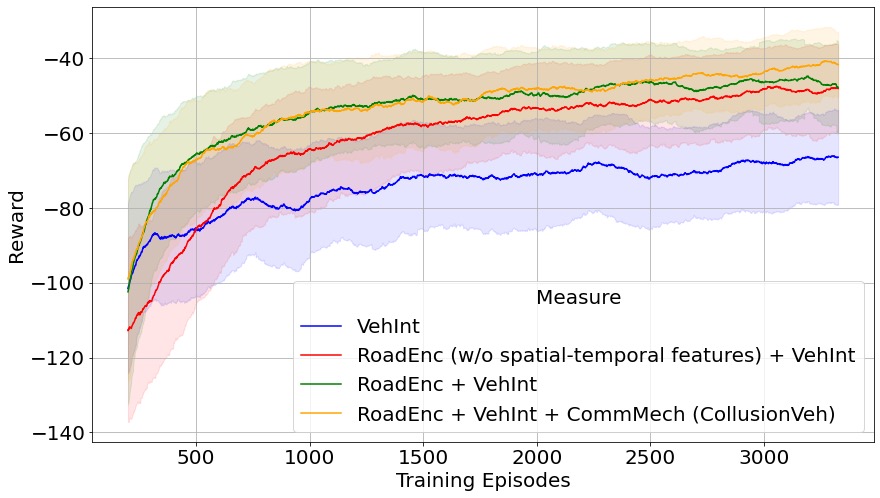}
    \caption{Training curves.}
    \label{fig:training}
\end{figure}
To study the effectiveness of each component in our framework, we conduct detailed ablation study considering the following three variants of \texttt{CollusionVeh}. \\
\textbf{\texttt{VehInt}}: In this setting, each colluding vehicle is treated as an independent agent. Specifically, we remove the parameter sharing mechanism and forbid communication between agents. In other words, this is equivalent to only applying an \texttt{VehInt} directly transforming raw observation into action. \\
\textbf{\texttt{RoadEnc} (w/o spatial-temporal features) + \texttt{VehInt}}: In this setting, agents do not have access to spatio-temporal embeddings and are not allowed to communicate with each other. \\
\textbf{\texttt{RoadEnc} + \texttt{VehInt}}: In this setting, we keep the first two components of \textsc{CollusionVeh} but remove \texttt{CommMech}. \\
The experiment results are shown in Table \ref{table:main_results} (row 5-8) and an example of the comparison between training curves for random seed 42 is shown in Fig. \ref{fig:training}. From table, we can see that all three metrics for colluding vehicles get improved as more components are included. From training curves, we see that the convergence takes longer when spatio-temporal features are discarded. Also, whether \texttt{RoadEnc} is adopted makes a big difference, demonstrating that the parameter sharing techniques for learning global embedding mechanisms are really helpful for agents to explore and understand the complex traffic environment efficiently. 
\subsection{Sensitivity Analysis}
\begin{figure}[h]
    \centering
    \includegraphics[width=.8\columnwidth]{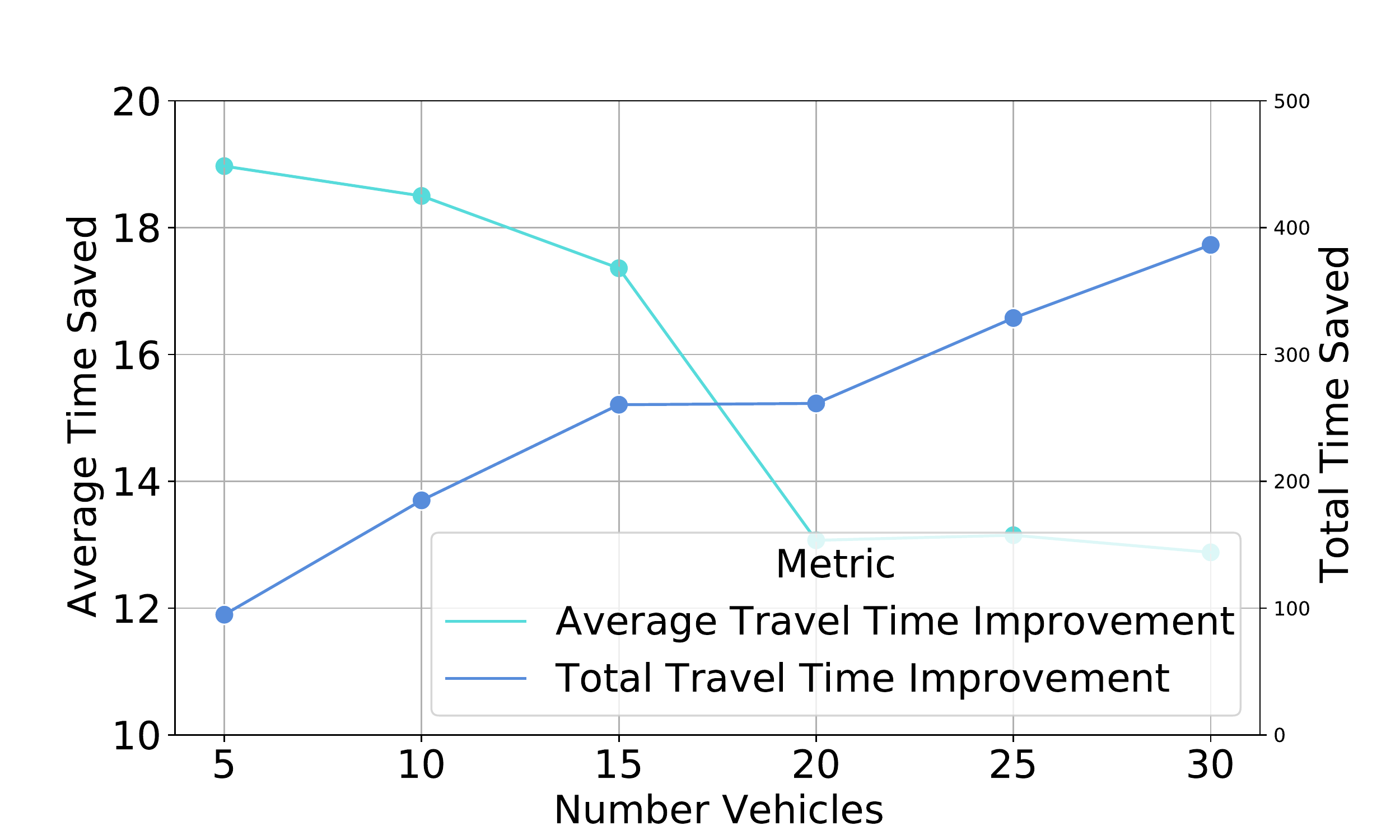}
    \caption{Travel time improvement with different number of agents.}
    \label{fig:sensitivity}
    \includegraphics[width=.8\columnwidth]{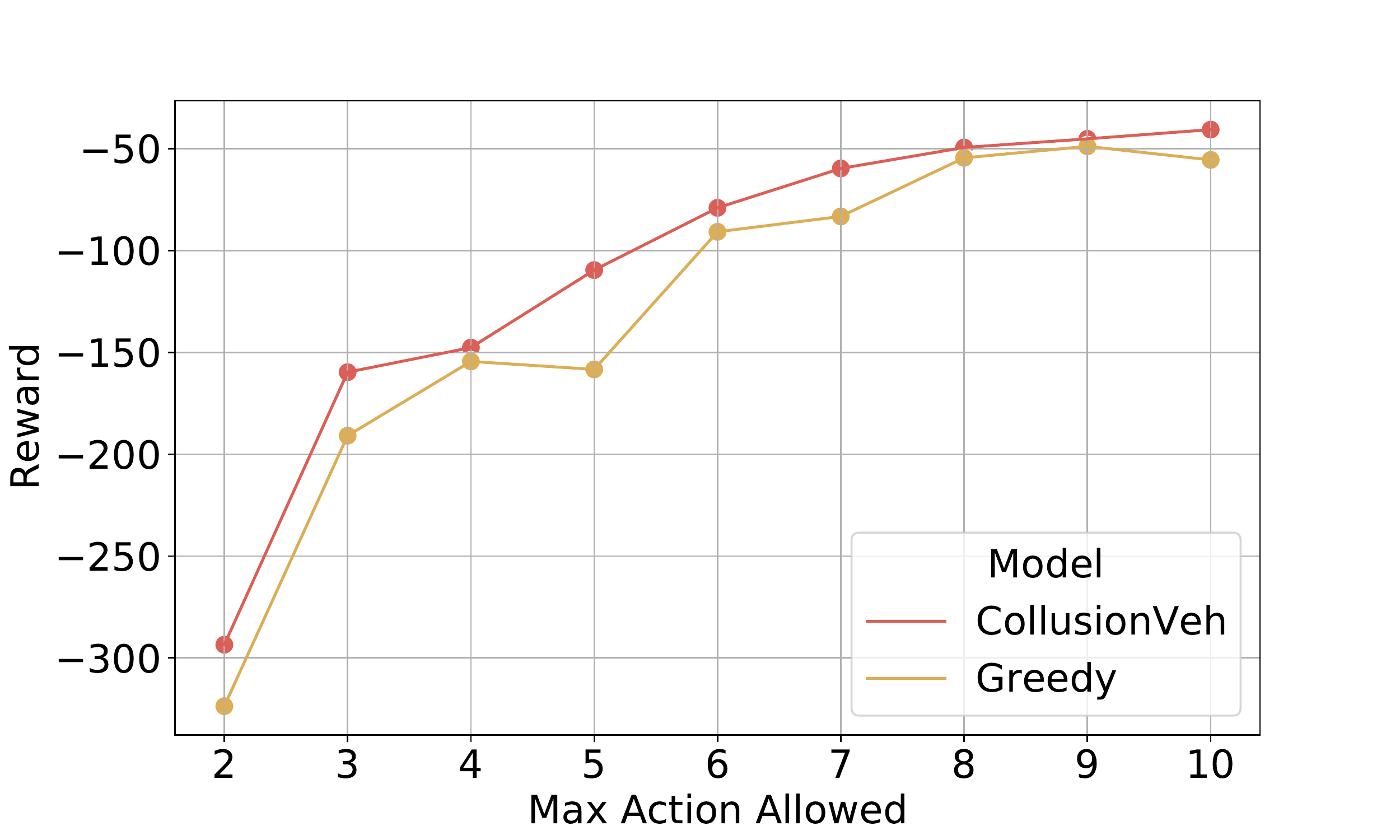}
    \caption{\texttt{CollusionVeh} vs Greedy strategy with varying action spaces.}
    \label{fig:action}
\end{figure}
In practice, attackers would not always find the same size of collusion and achieve the same action space. Therefore, to understand the impact of collusion size and action space, we perform two sets of sensitivity analyses. The first analysis examines whether varying sizes of the collusion group will make a difference, considering collusions of 5, 10, 15, 20, 25, and 30 vehicles. The second analysis inspects the performance of \textsc{CollusionVeh} with maximum possible action from 2 to 10. To make results comparable, we use the same seed for all experiments in this section. In this way, for example, when 25 colluding vehicles are selected, they are a subset of the 30 vehicle group rather than another random sample of 25 vehicles. \\
Results for the first analysis are shown in Fig \ref{fig:sensitivity} where two lines indicating average time saved and total time saved, respectively. We see that as more vehicles take part in the collusion, although they eventually manage to save more total travel time, the average individual gain is compromised. Results for the first analysis are shown in Fig \ref{fig:action} where two lines indicate the cumulative rewards achieved by \textsc{CollusionVeh} and Greedy approach (always report the maximum possible action), respectively. We observe that as action space enlarges, greedy approach does not always improve the rewards but \texttt{CollusionVeh} constantly takes advantage of the greater flexibility to secure more rewards. Another observation is that a small action space can already result in significant improvements in rewards, which are defined to be strongly correlated with reduction in waiting time ({\em i.e.}, when action space \textbf{A} = $\{0, 1,2,3\}$, the cumulative reward is improved by ~65.7\%).
\subsection{Insights and Suggestions}
In the previous sections, we demonstrate that vehicles can effectively reduce their waiting time at intersections by forming collusion groups and attacking DRL-based ATCS with falsified information. The implications of these results are significant as plenty of resources are invested in developing connected traffic environment and the scenario assumed in this paper may come true in the near future. In order to prevent such collusion from happening in the real world, we provide the following suggestions based on our study:
\begin{itemize}
    \item A strict certification mechanism for connected vehicles is critical in order to fundamentally address this problem while other issues such as privacy concern may make this process challenging.
    \item Updating ATCS's policy frequently should help prevent them from being attacked. As we see in Figure \ref{fig:training}, for ATCS powered by a DRL model with elaborate design ({\em e.g.}, \textsc{MA2C}), it takes colluding vehicles a certain amount of time to learn the environment and launch successful attacks ({\em i.e.}, 1000 episodes of training may take at least a year in real life).
    \item From an algorithmic perspective, equipping ATCS with real-time anomaly detection mechanisms may enable immediate recognition of attacks and effective countermeasures taken in time. Moreover, robust DRL, as a popular research area, can mitigate the effects of the falsified information.
\end{itemize}

\section{Conclusion}
This paper for the first time formulates a novel and realistic problem that in the connected and intelligent traffic environment, vehicles can form into a collusion group and reduce their total travel time by ``cheating'' the adaptive traffic signal control systems with falsified information. We introduce \textsc{CollusionVeh}, a generic DRL-based framework for coordinating colluding vehicles' attacks to optimize their benefits ({\em i.e.}, time saved). Both global embeddings and communication mechanism designed in \textsc{CollusionVeh} are interpretable and contributive. Experimental results indicate that colluding is more effective than each vehicle greedily sending falsified information, and the colluding effect will decrease if the number of colluding vehicles increases. Several insights and suggestions are also discussed to protect systems from such attacks.

Many future research directions are worth exploring: 1) in real life, ATCS may integrate several types of data for decision making ({\em e.g.}, traffic volume, traffic speed, vehicle position, etc.). Therefore, we may consider a more diverse action space for colluding vehicles; 2) to fully test the generalizability of this model, experiments on different types DRL-based ATCS policies are necessary; 3) in real life, attacks can be costly despite the benefits they bring, how to include such cost in the attack model is a remaining challenge; 4) it is also worth investigating the corresponding defending mechanism to actively detect and filter out the falsified information sent by colluding vehicles.

\clearpage
\bibstyle{aaai22}
\bibliography{refs}

\begin{thebibliography}{40}
\providecommand{\natexlab}[1]{#1}

\bibitem[{Behzadan and Munir(2017{\natexlab{a}})}]{behzadan2017vulnerability}
Behzadan, V.; and Munir, A. 2017{\natexlab{a}}.
\newblock Vulnerability of Deep Reinforcement Learning to Policy Induction
  Attacks.
\newblock arXiv:1701.04143.

\bibitem[{Behzadan and
  Munir(2017{\natexlab{b}})}]{10.1007/978-3-319-62416-7_19}
Behzadan, V.; and Munir, A. 2017{\natexlab{b}}.
\newblock Vulnerability of Deep Reinforcement Learning to Policy Induction
  Attacks.
\newblock In Perner, P., ed., \emph{Machine Learning and Data Mining in Pattern
  Recognition}, 262--275. Cham: Springer International Publishing.
\newblock ISBN 978-3-319-62416-7.

\bibitem[{Behzadan and Munir(2018)}]{behzadan2018faults}
Behzadan, V.; and Munir, A. 2018.
\newblock The Faults in Our Pi Stars: Security Issues and Open Challenges in
  Deep Reinforcement Learning.
\newblock arXiv:1810.10369.

\bibitem[{Bellman(1957)}]{bellman1957markovian}
Bellman, R. 1957.
\newblock A Markovian decision process.
\newblock \emph{Journal of mathematics and mechanics}, 6(5): 679--684.

\bibitem[{Casas(2017)}]{casas2017deep}
Casas, N. 2017.
\newblock Deep Deterministic Policy Gradient for Urban Traffic Light Control.
\newblock arXiv:1703.09035.

\bibitem[{Chen et~al.(2020)Chen, Wei, Xu, Zheng, Yang, Xiong, Xu, and
  Li}]{Chen_Wei_Xu_Zheng_Yang_Xiong_Xu_Li_2020}
Chen, C.; Wei, H.; Xu, N.; Zheng, G.; Yang, M.; Xiong, Y.; Xu, K.; and Li, Z.
  2020.
\newblock Toward A Thousand Lights: Decentralized Deep Reinforcement Learning
  for Large-Scale Traffic Signal Control.
\newblock \emph{Proceedings of the AAAI Conference on Artificial Intelligence},
  34(04): 3414--3421.

\bibitem[{Chen et~al.(2018)Chen, Yin, Feng, Mao, and Liu}]{Chen2018ExposingCA}
Chen, Q.; Yin, Y.; Feng, Y.; Mao, Z.~M.; and Liu, H.~X. 2018.
\newblock Exposing Congestion Attack on Emerging Connected Vehicle based
  Traffic Signal Control.
\newblock In \emph{NDSS}.

\bibitem[{Chen et~al.(2021)Chen, Dong, Ha, Li, and Labi}]{chen2021graph}
Chen, S.; Dong, J.; Ha, P.; Li, Y.; and Labi, S. 2021.
\newblock Graph neural network and reinforcement learning for multi-agent
  cooperative control of connected autonomous vehicles.
\newblock \emph{Computer-Aided Civil and Infrastructure Engineering}, 36(7):
  838--857.

\bibitem[{Chu et~al.(2019)Chu, Wang, Codec{\`a}, and Li}]{chu2019multi}
Chu, T.; Wang, J.; Codec{\`a}, L.; and Li, Z. 2019.
\newblock Multi-Agent Deep Reinforcement Learning for Large-Scale Traffic
  Signal Control.
\newblock \emph{IEEE Transactions on Intelligent Transportation Systems}.

\bibitem[{CLTC(2021)}]{UCBerkelyReport}
CLTC, U. 2021.
\newblock The Cybersecurity Risks of Smart City Technologies: What Do The
  Experts Think?

\bibitem[{Codeca and Harri(2018)}]{SUMO2018:Monaco_SUMO_Traffic_MoST}
Codeca, L.; and Harri, J. 2018.
\newblock Monaco SUMO Traffic (MoST) Scenario: A 3D Mobility Scenario for
  Cooperative ITS.
\newblock In Wie\{\textbackslash{}ss\}ner, E.; L\textbackslash{}"ucken, L.;
  Hilbrich, R.; Fl\textbackslash{}"otter\textbackslash{}"od, Y.-P.; Erdmann,
  J.; Bieker-Walz, L.; and Behrisch, M., eds., \emph{SUMO 2018- Simulating
  Autonomous and Intermodal Transport Systems}, volume~2 of \emph{EPiC Series
  in Engineering}, 43--55. EasyChair.

\bibitem[{{Cranley, Ellen}(2020)}]{cranley_2020}
{Cranley, Ellen}. 2020.
\newblock 8 cities that have been crippled by CYBERATTACKS - and what they did
  to fight them.

\bibitem[{Dosovitskiy et~al.(2017)Dosovitskiy, Ros, Codevilla, Lopez, and
  Koltun}]{pmlr-v78-dosovitskiy17a}
Dosovitskiy, A.; Ros, G.; Codevilla, F.; Lopez, A.; and Koltun, V. 2017.
\newblock {CARLA}: {An} Open Urban Driving Simulator.
\newblock In Levine, S.; Vanhoucke, V.; and Goldberg, K., eds.,
  \emph{Proceedings of the 1st Annual Conference on Robot Learning}, volume~78
  of \emph{Proceedings of Machine Learning Research}, 1--16. PMLR.

\bibitem[{Feng et~al.(2018)Feng, Huang, Chen, Liu, and
  Mao}]{doi:10.1177/0361198118756885}
Feng, Y.; Huang, S.; Chen, Q.~A.; Liu, H.~X.; and Mao, Z.~M. 2018.
\newblock Vulnerability of Traffic Control System Under Cyberattacks with
  Falsified Data.
\newblock \emph{Transportation Research Record}, 2672(1): 1--11.

\bibitem[{{Freed, Benjamin}(2021)}]{freed_2021}
{Freed, Benjamin}. 2021.
\newblock Atlanta's ransomware attack Destroyed years of police dashboard
  camera footage.

\bibitem[{Gleave et~al.(2019)Gleave, Dennis, Kant, Wild, Levine, and
  Russell}]{DBLP:journals/corr/abs-1905-10615}
Gleave, A.; Dennis, M.; Kant, N.; Wild, C.; Levine, S.; and Russell, S. 2019.
\newblock Adversarial Policies: Attacking Deep Reinforcement Learning.
\newblock \emph{CoRR}, abs/1905.10615.

\bibitem[{Gong et~al.(2019)Gong, Abdel-Aty, Cai, and Rahman}]{GONG2019100020}
Gong, Y.; Abdel-Aty, M.; Cai, Q.; and Rahman, M.~S. 2019.
\newblock Decentralized network level adaptive signal control by multi-agent
  deep reinforcement learning.
\newblock \emph{Transportation Research Interdisciplinary Perspectives}, 1:
  100020.

\bibitem[{Gunarathna et~al.(2019)Gunarathna, Xie, Tanin, Karunasekara, and
  Borovica-Gajic}]{gunarathna2019dynamic}
Gunarathna, U.; Xie, H.; Tanin, E.; Karunasekara, S.; and Borovica-Gajic, R.
  2019.
\newblock Dynamic graph configuration with reinforcement learning for connected
  autonomous vehicle trajectories.
\newblock \emph{arXiv preprint arXiv:1910.06788}.

\bibitem[{Ilahi et~al.(2021)Ilahi, Usama, Qadir, Janjua, Al-Fuqaha, Hoang, and
  Niyato}]{ilahi2021challenges}
Ilahi, I.; Usama, M.; Qadir, J.; Janjua, M.~U.; Al-Fuqaha, A.; Hoang, D.~T.;
  and Niyato, D. 2021.
\newblock Challenges and Countermeasures for Adversarial Attacks on Deep
  Reinforcement Learning.
\newblock arXiv:2001.09684.

\bibitem[{Isaac et~al.(2008)Isaac, Camara, Zeadally, and
  Marquez}]{isaac2008secure}
Isaac, J.~T.; Camara, J.~S.; Zeadally, S.; and Marquez, J.~T. 2008.
\newblock A secure vehicle-to-roadside communication payment protocol in
  vehicular ad hoc networks.
\newblock \emph{Computer Communications}, 31(10): 2478--2484.

\bibitem[{KDD(2021)}]{city_brain_challenge}
KDD. 2021.
\newblock City Brain Challenge.

\bibitem[{Khan et~al.(2017)Khan, Ahmad, Cao, Jalal, Asif, ul~Haq, and
  Cruichshank}]{khan2017certificate}
Khan, T.; Ahmad, N.; Cao, Y.; Jalal, S.~A.; Asif, M.; ul~Haq, S.; and
  Cruichshank, H. 2017.
\newblock Certificate revocation in vehicular ad hoc networks techniques and
  protocols: a survey.
\newblock \emph{Science China Information Sciences}, 60(10): 1--18.

\bibitem[{Mnih et~al.(2016)Mnih, Badia, Mirza, Graves, Lillicrap, Harley,
  Silver, and Kavukcuoglu}]{pmlr-v48-mniha16}
Mnih, V.; Badia, A.~P.; Mirza, M.; Graves, A.; Lillicrap, T.; Harley, T.;
  Silver, D.; and Kavukcuoglu, K. 2016.
\newblock Asynchronous Methods for Deep Reinforcement Learning.
\newblock In Balcan, M.~F.; and Weinberger, K.~Q., eds., \emph{Proceedings of
  The 33rd International Conference on Machine Learning}, volume~48 of
  \emph{Proceedings of Machine Learning Research}, 1928--1937. New York, New
  York, USA: PMLR.

\bibitem[{Mousavi et~al.(2017)Mousavi, Schukat, Corcoran, and
  Howley}]{DBLP:journals/corr/MousaviS0H17}
Mousavi, S.~S.; Schukat, M.; Corcoran, P.; and Howley, E. 2017.
\newblock Traffic Light Control Using Deep Policy-Gradient and Value-Function
  Based Reinforcement Learning.
\newblock \emph{CoRR}, abs/1704.08883.

\bibitem[{Nishi et~al.(2018)Nishi, Otaki, Hayakawa, and Yoshimura}]{8569301}
Nishi, T.; Otaki, K.; Hayakawa, K.; and Yoshimura, T. 2018.
\newblock Traffic Signal Control Based on Reinforcement Learning with Graph
  Convolutional Neural Nets.
\newblock In \emph{2018 21st International Conference on Intelligent
  Transportation Systems (ITSC)}, 877--883.

\bibitem[{Prashanth and Bhatnagar(2011)}]{6082823}
Prashanth, L.~A.; and Bhatnagar, S. 2011.
\newblock Reinforcement learning with average cost for adaptive control of
  traffic lights at intersections.
\newblock In \emph{2011 14th International IEEE Conference on Intelligent
  Transportation Systems (ITSC)}, 1640--1645.

\bibitem[{Schulman et~al.(2017)Schulman, Wolski, Dhariwal, Radford, and
  Klimov}]{schulman2017proximal}
Schulman, J.; Wolski, F.; Dhariwal, P.; Radford, A.; and Klimov, O. 2017.
\newblock Proximal policy optimization algorithms.
\newblock \emph{arXiv preprint arXiv:1707.06347}.

\bibitem[{Sutton and Barto(2018)}]{sutton2018reinforcement}
Sutton, R.~S.; and Barto, A.~G. 2018.
\newblock \emph{Reinforcement learning: An introduction}.
\newblock MIT press.

\bibitem[{Toh(2001)}]{toh2001ad}
Toh, C.~K. 2001.
\newblock \emph{Ad hoc mobile wireless networks: protocols and systems}.
\newblock Pearson Education.

\bibitem[{USAGov(2021)}]{the_white_house_2021}
USAGov. 2021.
\newblock FACT sheet: President Biden announces support for the Bipartisan
  infrastructure framework.

\bibitem[{Wang et~al.(2020{\natexlab{a}})Wang, Shi, Wu, Miranda-Moreno, and
  Sun}]{wangmulti}
Wang, J.; Shi, T.; Wu, Y.; Miranda-Moreno, L.; and Sun, L. 2020{\natexlab{a}}.
\newblock Multi-agent Graph Reinforcement Learning for Connected Automated
  Driving.
\newblock \emph{Conference: ICML Workshop on AI for Autonomous Driving}.

\bibitem[{Wang et~al.(2020{\natexlab{b}})Wang, Xu, Niu, Tan, Chen, and
  Xiong}]{wang2020stmarl}
Wang, Y.; Xu, T.; Niu, X.; Tan, C.; Chen, E.; and Xiong, H. 2020{\natexlab{b}}.
\newblock STMARL: A Spatio-Temporal Multi-Agent Reinforcement Learning Approach
  for Cooperative Traffic Light Control.
\newblock arXiv:1908.10577.

\bibitem[{Wei et~al.(2019)Wei, Xu, Zhang, Zheng, Zang, Chen, Zhang, Zhu, Xu,
  and Li}]{Wei_2019}
Wei, H.; Xu, N.; Zhang, H.; Zheng, G.; Zang, X.; Chen, C.; Zhang, W.; Zhu, Y.;
  Xu, K.; and Li, Z. 2019.
\newblock CoLight.
\newblock \emph{Proceedings of the 28th ACM International Conference on
  Information and Knowledge Management}.

\bibitem[{Wei et~al.(2018)Wei, Zheng, Yao, and Li}]{10.1145/3219819.3220096}
Wei, H.; Zheng, G.; Yao, H.; and Li, Z. 2018.
\newblock IntelliLight: A Reinforcement Learning Approach for Intelligent
  Traffic Light Control.
\newblock In \emph{Proceedings of the 24th ACM SIGKDD International Conference
  on Knowledge Discovery \& Data Mining}, KDD '18, 2496–2505. New York, NY,
  USA: Association for Computing Machinery.
\newblock ISBN 9781450355520.

\bibitem[{Xu et~al.(2021)Xu, Wang, Wang, Jia, and
  Lu}]{Xu_Wang_Wang_Jia_Lu_2021}
Xu, B.; Wang, Y.; Wang, Z.; Jia, H.; and Lu, Z. 2021.
\newblock Hierarchically and Cooperatively Learning Traffic Signal Control.
\newblock \emph{Proceedings of the AAAI Conference on Artificial Intelligence},
  35(1): 669--677.

\bibitem[{Zhang et~al.(2021)Zhang, Chen, Boning, and
  Hsieh}]{DBLP:journals/corr/abs-2101-08452}
Zhang, H.; Chen, H.; Boning, D.~S.; and Hsieh, C. 2021.
\newblock Robust Reinforcement Learning on State Observations with Learned
  Optimal Adversary.
\newblock \emph{CoRR}, abs/2101.08452.

\bibitem[{Zhang et~al.(2020)Zhang, Chen, Xiao, Li, Boning, and
  Hsieh}]{DBLP:journals/corr/abs-2003-08938}
Zhang, H.; Chen, H.; Xiao, C.; Li, B.; Boning, D.~S.; and Hsieh, C. 2020.
\newblock Robust Deep Reinforcement Learning against Adversarial Perturbations
  on Observations.
\newblock \emph{CoRR}, abs/2003.08938.

\bibitem[{Zhang, Yang, and Zha(2020)}]{10.5555/3398761.3399082}
Zhang, Z.; Yang, J.; and Zha, H. 2020.
\newblock Integrating Independent and Centralized Multi-Agent Reinforcement
  Learning for Traffic Signal Network Optimization.
\newblock In \emph{Proceedings of the 19th International Conference on
  Autonomous Agents and MultiAgent Systems}, AAMAS '20, 2083–2085. Richland,
  SC: International Foundation for Autonomous Agents and Multiagent Systems.
\newblock ISBN 9781450375184.

\bibitem[{Zheng et~al.(2019)Zheng, Xiong, Zang, Feng, Wei, Zhang, Li, Xu, and
  Li}]{DBLP:journals/corr/abs-1905-04722}
Zheng, G.; Xiong, Y.; Zang, X.; Feng, J.; Wei, H.; Zhang, H.; Li, Y.; Xu, K.;
  and Li, Z. 2019.
\newblock Learning Phase Competition for Traffic Signal Control.
\newblock \emph{CoRR}, abs/1905.04722.

\bibitem[{ZMEScience(2021)}]{zme_science_2021}
ZMEScience. 2021.
\newblock A{I} Traffic Management Could Finally Declog Urban Roads.

\end{thebibliography}

\clearpage
\section{Appendix}

\setcounter{table}{0}
\renewcommand{\thetable}{A\arabic{table}}
\setcounter{figure}{0}
\renewcommand{\thefigure}{A\arabic{figure}}
\subsection{Road Network}
\begin{figure}[h!]
    \centering
    \includegraphics[width=1\columnwidth]{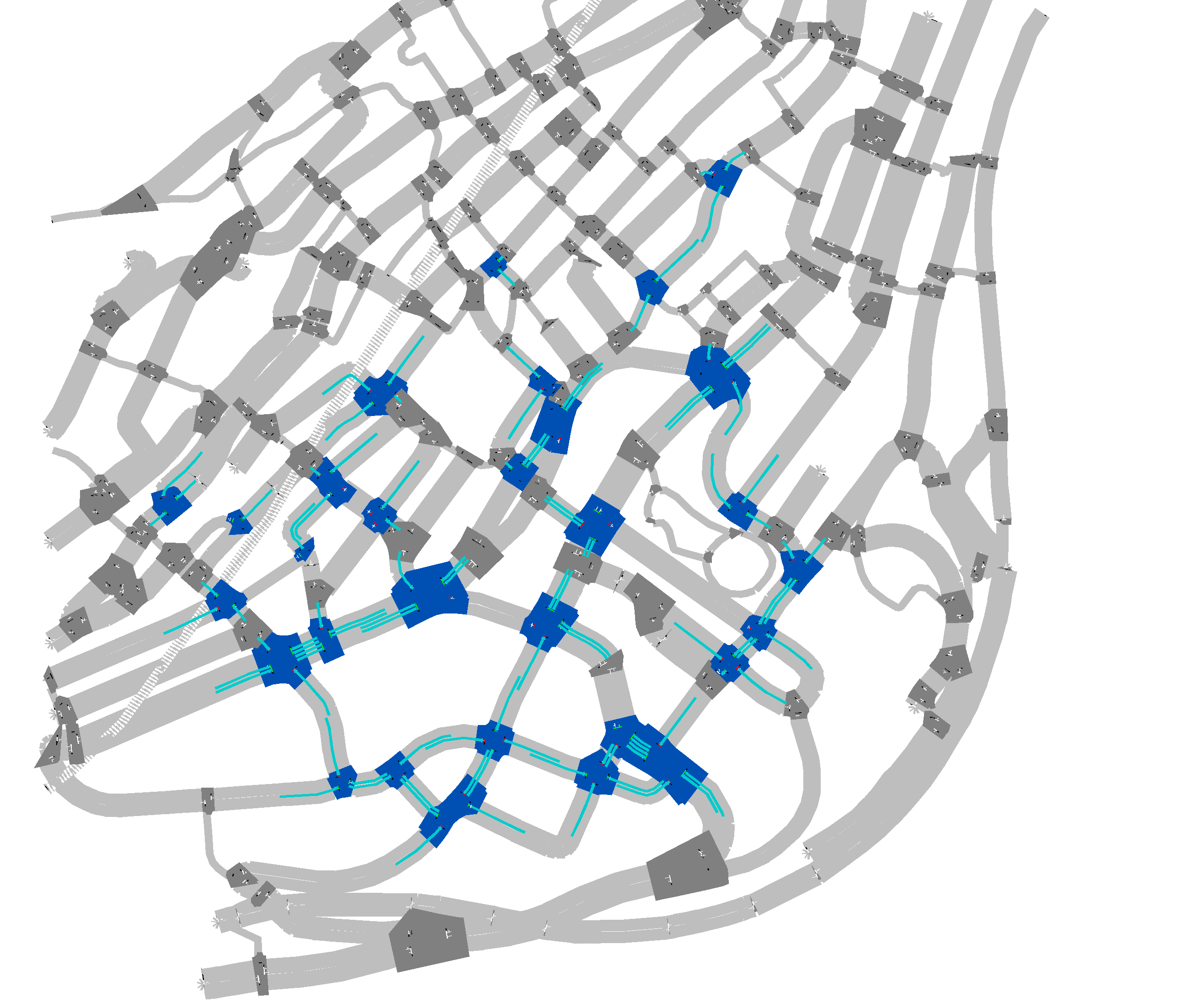}
    \caption{Monaco City Road Network}
    \label{fig:road_network}
\end{figure}

The traffic network used in our experiments is adapted from the real road network in Monaco City by \cite{SUMO2018:Monaco_SUMO_Traffic_MoST}. An visualization is shown in Fig \ref{fig:road_network}.

\subsection{Full Results}
In this section, we show the complete experiment results for the table and figures presented in the main part. Table \ref{table:full_main_results} contains complete results for all 5 experiments for comparing models and ablation studies. Table \ref{table:full_sensitivity} contains complete results for experiments with varying number of colluding vehicles. Table \ref{table:full_action} contains complete results for experiments with varying action space. In Table \ref{table:full_action}, \textsc{Greedy} is the policy that makes agent always report the maximum action allowed. For average colluding vehicle travel time and waiting time, we highlight the lowest value for each comparison. For average other vehicle travel time and waiting time, we highlight the highest value which means more time being consumed. From these tables, we can see that \textsc{CollusionVeh} constantly achieves the best results across reward, average collusion travel time, and average collusion waiting time but it does not necessarily cause the most significant increase in average other vehicle travel time and waiting time. In other words, although \textsc{CollusionVeh} is guaranteed to save time for colluding vehicles, it does not always cause the greatest "harm" for other vehicles.
\begin{table*}[b!]
\caption{Full Results for Table \ref{table:main_results}}
\label{table:full_main_results}
\resizebox{\textwidth}{!}{%
\begin{tabular}{p{2cm}p{5cm}P{1.5cm}P{1.5cm}P{1.5cm}P{1.5cm}P{1.5cm}P{3cm}P{3cm}P{3cm}}
\hline
Random Seed         & Metric                        & \textsc{All One}  & \textsc{All Five} & \textsc{All Ten} & \textsc{Random}  & \texttt{VehInt} & \texttt{RoadEnc}(reduced) + \texttt{VehInt} & \texttt{RoadEnc} + \texttt{VehInt} & \textsc{CollusionVeh} \\ \hline
\multirow{5}{*}{0}  & Reward                        & -495.45 & -140.82 & -64.41 & -143.62 & -63.35 & -44.84 & -42.52 & \textbf{-41.51} \\ 
                    & Collusion Travel Time Avg  & 83.86   & 72.45   & 69.86  & 71.30   & 69.59  & 69.58  & 69.25  & \textbf{68.66}  \\ 
                    & Collusion Waiting Time Avg & 15.83   & 4.50    & 1.97   & 3.40    & 1.67   & 1.60   & 1.27   & \textbf{0.77}   \\ 
                    & Other Travel Time Avg      & 121.68  & 118.01  & 126.87 & 118.52  & \textbf{141.57} & 136.59 & 120.28 & 130.82 \\
                    & Other Waiting Time Avg     & 25.28   & 22.03   & 30.46  & 22.36   & \textbf{43.81}  & 39.73  & 24.37  & 34.88  \\ \hline
\multirow{5}{*}{1}  & Reward                        & -620.97 & -133.45 & -37.38 & -111.37 & -36.04 & -36.55 & -31.73 & \textbf{-30.16} \\
                    & Collusion Travel Time Avg  & 85.53   & 69.77   & 66.81  & 69.09   & 66.85  & 66.65  & \textbf{66.54}  & 66.56  \\
                    & Collusion Waiting Time Avg & 20.00   & 4.23    & 1.20   & 3.60    & 1.30   & 1.17   & \textbf{1.00}   & \textbf{1.00}   \\
                    & Other Travel Time Avg      & 121.65  & 137.72  & 135.09 & 134.71  & 115.05 & 133.98 & 141.07 & \textbf{142.16} \\
                    & Other Waiting Time Avg     & 25.01   & 40.83   & 38.60  & 37.52   & 18.69  & 36.77  & 43.15  & \textbf{44.61}  \\ \hline
\multirow{5}{*}{10} & Reward                        & -378.95 & -159.34 & -71.88 & -113.34 & -65.13 & -65.07 & -60.83 & \textbf{-49.55} \\
                    & Collusion Travel Time Avg  & 81.50   & 74.42   & 71.78  & 72.65   & 71.54  & 72.63  & \textbf{70.87}  & 70.96  \\
                    & Collusion Waiting Time Avg & 11.87   & 5.03    & 2.30   & 3.07    & 2.17   & 2.97   & \textbf{1.43}   & 1.47   \\ 
                    & Other Travel Time Avg      & 122.03  & 119.76  & \textbf{154.86} & 134.11  & 113.75 & 112.18 & 118.12 & 149.70 \\
                    & Other Waiting Time Avg     & 25.52   & 24.13   & \textbf{55.41}  & 38.36   & 18.17  & 17.30  & 22.01  & 51.76  \\ \hline
\multirow{5}{*}{12} & Reward                        & -450.63 & -144.08 & -80.64 & -125.44 & -34.69 & -33.32 & -29.63 & \textbf{-27.08} \\ 
& Collusion Travel Time Avg  & 80.95   & 71.25   & 69.50  & 68.82   & 68.18  & 68.02  & 68.04  & \textbf{67.96}  \\
                    & Collusion Waiting Time Avg & 14.13   & 4.37    & 2.60   & 2.00    & 1.37   & 1.19   & \textbf{1.17}   & \textbf{1.17}   \\  & Other Travel Time Avg      & 122.10  & 135.33  & 138.70 & 127.67  & 123.07 & 125.49 & 145.25 & \textbf{149.83} \\   & Other Waiting Time Avg     & 25.38   & 38.37   & 42.18  & 31.66   & 26.15  & 30.00  & 48.30  & \textbf{51.40}  \\ \hline
\multirow{5}{*}{42} & Reward                        & -464.02 & -158.35 & -55.56 & -146.31 & -67.26 & -48.83 & -44.99 & \textbf{-40.70} \\
                    & Collusion Travel Time Avg  & 84.48   & 74.56   & 71.24  & 73.11   & 71.61  & 70.85  & 71.61  & \textbf{70.60}  \\     & Collusion Waiting Time Avg & 14.83   & 4.87    & 1.67   & 3.47    & 1.90   & 1.03   & 1.83   & \textbf{0.87}   \\    & Other Travel Time Avg      & 121.72  & 145.18  & 117.37 & 140.18  & \textbf{156.63} & 118.08 & 138.29 & 118.00 \\   & Other Waiting Time Avg     & 25.34   & 47.19   & 21.32  & 43.55   & \textbf{57.45}  & 22.36  & 41.18  & 22.01  \\ \hline
\end{tabular}%
}
\end{table*}

\begin{table*}[t!]
\caption{Full Results for Fig \ref{fig:sensitivity}}
\label{table:full_sensitivity}
\resizebox{\textwidth}{!}{%
\begin{tabular}{llllllll}
\hline
Metric                                  & Mode  & 5               & 10              & 15              & 20             & 25              & 30              \\ \hline
\multirow{2}{*}{Reward}                     & \textsc{Collusion Veh} & \textbf{-4.14} & \textbf{-6.6} & \textbf{-16.72} & \textbf{-14.15} & \textbf{-43.31} & \textbf{-40.7} \\ 
                                        & \textsc{All One}  & -100            & -191.33         & -283            & -293.77        & -392.3          & -464.02         \\ \hline
\multirow{2}{*}{Collusion Travel Time Avg}  & \textsc{CollusionVeh} & \textbf{69.04} & \textbf{69.7} & \textbf{69.22}  & \textbf{67.36}  & \textbf{70.58}  & \textbf{70.6}  \\ 
                                        & \textsc{All One}  & 88.01           & 88.2            & 86.58           & 80.43          & 83.73           & 84.48           \\ \hline
\multirow{2}{*}{Collusion Waiting Time Avg} & \textsc{CollusionVeh} & \textbf{1}     & \textbf{0.2}  & \textbf{1}      & \textbf{1.05}   & \textbf{1.88}   & \textbf{0.87}  \\
                                        & \textsc{All One}  & 20              & 18.8            & 18.27           & 14.15          & 15.16           & 14.83           \\ \hline
\multirow{2}{*}{Other Travel Time Avg}  & \textsc{CollusionVeh} & 115.75          & \textbf{140.53} & \textbf{125.73} & \textbf{127.4} & \textbf{127.46} & 118             \\ 
                                        & \textsc{All One}  & \textbf{120.96} & 121.04          & 121.21          & 121.59         & 121.59          & \textbf{121.72} \\ \hline
\multirow{2}{*}{Other Waiting Time Avg} & \textsc{CollusionVeh} & 19.88           & \textbf{43.23}  & \textbf{29.45}  & \textbf{30.8}  & \textbf{31.81}  & 22.01           \\ 
                                        & \textsc{All One}  & \textbf{24.77}  & 24.84           & 24.91           & 25.15          & 25.21           & \textbf{25.34}  \\ \hline
\end{tabular}%
}
\end{table*}

\begin{table*}[t!]
\caption{Full Results for Fig \ref{fig:action}}
\label{table:full_action}
\resizebox{\textwidth}{!}{%
\begin{tabular}{lllllllllll}
\hline
Metric &
  Mode &
  2 &
  3 &
  4 &
  5 &
  6 &
  7 &
  8 &
  9 &
  10 \\ \hline
\multirow{2}{*}{Reward} &
  \textsc{CollusionVeh} &
  \textbf{-293.52} &
  \textbf{-159.75} &
  \textbf{-147.67} &
  \textbf{-109.63} &
  \textbf{-79.09} &
  \textbf{-59.76} &
  \textbf{-49.46} &
  \textbf{-45.23} &
  \textbf{-40.70} \\  
 &
  Geedy &
  -323.69 &
  -190.92 &
  -154.48 &
  -158.39 &
  -90.86 &
  -83.30 &
  -54.56 &
  -48.90 &
  -55.56 \\ \hline
\multirow{2}{*}{Collusion Travel Time Avg} &
  \textsc{CollusionVeh} &
  \textbf{80.13} &
  \textbf{75.92} &
  \textbf{74.75} &
  \textbf{74.56} &
  \textbf{72.40} &
  \textbf{72.05} &
  \textbf{71.34} &
  \textbf{71.20} &
  \textbf{70.60} \\ 
 &
 \textsc{Greedy}&
  84.46 &
  79.23 &
  75.63 &
  74.58 &
  73.00 &
  72.35 &
  71.45 &
  71.40 &
  71.24 \\ \hline
\multirow{2}{*}{Collusion Waiting Time Avg} &
  \textsc{CollusionVeh} &
  \textbf{10.37} &
  \textbf{6.07} &
  \textbf{4.97} &
  \textbf{4.87} &
  \textbf{2.93} &
  \textbf{2.40} &
  \textbf{1.70} &
  \textbf{1.50} &
  \textbf{0.87} \\ 
 &
 \textsc{Greedy}&
  15.07 &
  9.53 &
  5.91 &
  \textbf{4.87} &
  3.56 &
  2.67 &
  1.73 &
  1.67 &
  1.67 \\ \hline
\multirow{2}{*}{Other Travel Time Avg} &
  \textsc{CollusionVeh} &
  135.16 &
  118.37 &
  116.74 &
  \textbf{145.18} &
  \textbf{125.97} &
  118.14 &
  116.15 &
  124.54 &
  \textbf{118.00} \\  
 &
 \textsc{Greedy}&
  \textbf{137.65} &
  \textbf{121.53} &
  \textbf{147.90} &
  114.72 &
  122.34 &
  \textbf{134.14} &
  \textbf{128.14} &
  \textbf{131.86} &
  117.37 \\ \hline
\multirow{2}{*}{Other Waiting Time Avg} &
  \textsc{CollusionVeh} &
  38.79 &
  22.54 &
  21.02 &
  \textbf{47.19} &
  \textbf{30.11} &
  22.07 &
  20.57 &
  28.86 &
  \textbf{22.01} \\ 
 &
 \textsc{Greedy}&
  \textbf{40.70} &
  \textbf{25.18} &
  \textbf{50.45} &
  19.30 &
  27.42 &
  \textbf{37.62} &
  \textbf{32.56} &
  \textbf{35.36} &
  21.32 \\ \hline
\end{tabular}%
}
\end{table*}

\subsection{Reproducibility}
In this section, to address the reproducibility, we provide the specifics about computing hardware and software used in our experiments. \\
All models with learnable parameters are trained on Amazon Elastic Compute Cloud(Amazon EC2), a part of Amazon Web Services(AWS) that allows users to rent virtual computers. We use two instances of c5.18xlarge type which provides 72 virtual CPUs and 144 GB memory. The training is done using rllib, an open-sourced reinforcement learning python library that provides easy-to-use and highly customizable training API.\\
For the core modules in \textsc{CollusionVeh}, \texttt{RoadEnc} contains an embedding mechanism with output size 16 for each type of features considered ({\em e.g. time, location, vehicles, colluding vehicles.}). \texttt{VehInt} has output size 64 and input size 64. \texttt{CommMech} takes an arbitrary number of neighboring policies and has output size 16. 
During training, the random seeds for sampling 5 groups of colluding vehicles are 0,1,10,12,and 42. These are chosen as five of the most popular random seeds. Since the goal of this paper is not to achieve the best performance by fine-tuning models, we keep most of the default hyperparameters provided by rllib with necessary adjustments only for speeding up the process.  
\end{document}